# Reaserchnig the Development of the Electrical Power System Using Systemically Evolutionary Algorithm

[1]Jerzy Tchorzewski, [2]Emil Chyzy

*Department of Artificial Intelligence,Faculty of Science,*

*University of Life Sciences and Humanities, Siedlce, Poland,*

[1]*Member of IEEE, jtchorzewski@interia.pl, Jerzy.Tchorzewski@uph.edu.pl*

[2]*Member of GENBIT Student Branch, emil.chyzy@gmail.com*

***Abstract.*** **The paper contains the concept and the results of research concerning the evolutionary algorithm, identified based on the systems control theory, which was called the Systemically of Evolutionary Algorithm (SAE). Special attention was paid to two elements of evolutionary algorithms, which have not been fully solved yet, i.e. to the methods used to create the initial population and the method of creating the robustness (fitness) function. Other elements of the SEA algorithm, i.a. cross-over, mutation, selection, etc. were also defined from a systemic point of view. Computational experiments were conducted using a selected subsystem of the Polish Electrical Power System and three programming languages: Java, $C^{++}$ and Matlab. Selected comparative results for the SAE algorithm in different implementations were also presented.**

## 1. Introduction to Systemically Evolutionary Algorithm

*In order to undertake research on the development of large systems of this class as the power system in terms of unmanned manufactory id necessary to obtain a model of the system, which as a great system it is possible to obtain only by way of identification. However, the models obtained by identification even with very accurate algorithms are burdened with large uncertainty. It is possible to improve the model, including using evolutionary algorithms, but oriented to needs arising from the development system. In an attempt to resolve this problem is proposed Systemically Evolutionary Algorithm[1] (EAS), which is implemented in several programming environments in order to choose the best environment for the development of power system design. In the present paper shows the some results of the implementation language for the environment: Java,* $C^{++}$ *and Matlab.*

*The study, published in this paper are a continuation of a broader research that leads Jerzy Tchorzewski since 1986, working to find the development model of the power system as a system of technical-economic developing. The results of previous studies, including The Systemically Evolutionary Algorithm has been published among others in [17-24]. In the literature there are no published studies of this type, hence it is not possible to review the literature. However the inspiration to study were among other works in the field of control theory and systems authors such as: Jozef Konieczny [9], Robert Staniszewski[2] [15], Ryszard Tadeusiewicz [16] and Slawomir T. Wierzchon [25].*

*Formulation of the concept of Systemically Evolutionary Algorithm (SAE) was based on the systems and control theory [7, 14, 16] in relation to the state of knowledge as regards evolutionary algorithms [1, 3-6, 9, 11-13, 15, 20] and the problem was described in the paper [17-19, 21-25]. The following*

---

[1] This paper is a part of wider research carried out by Dr. Ing. Jerzy Tchorzewski since 2011, which was published among others in [17-24]. At the moment author is working on the final version of his monograph, which intends to publish in Scientific & Academic Publishing, by which he was invited to write extended paper from the Conference 2011 UKSim 13th International Conference on Computer Modeling and Simulation (UKSim).

[2] Dr Jerzy Tchorzewski finished this book after the death of Professor Robert Staniszewski

*basic elements of the algorithms were defined such as system initial population generated directly from the system [21-25], systemically cross-over and mutation operators as operators defined based on the theory of operation systems, and in particular, using the notion of the operator of longitudinal-transverse sub-system [8, 17-19], system method of selection (system method of heuristic elimination) and system adaptation function defined based on the concept of fitness function, which was related to system discrepancies [14, 17-19].*

## 1.1. Operator of longitudinal-transverse subsysteming

The notion of subsysteming operator was introduced by Jozef Konieczny, who distinguished the operator of longitudinal subsysteming and transverse subsysteming [8]. The operator of longitudinal subsysteming divides the system into two subsystems: the control subsystem and the realization subsystem and the transverse subsysteming divides the system into the sub-system securing the functioning and operating subsystem, which may be written down as follows:

$$S_i \perp \left\langle S_i^s \middle| S_i^r \right\rangle \& / or \left\langle S_i^z \middle| S_i^o \right\rangle, \tag{1}$$

where:
$\perp$ - operator of longitudinal-transverse subsysteming,
$S_i^j$ - the sub-system of the $i^{th}$ longitudinal subsysteming and $j^{th}$ transverse subsysteming.

As a result of subsysteming, subsystems appear connected with each others by the streams that make them constitute a whole. The subsysteming operator understood in this way, as a model of development is connected with the integrated notation of knowledge about the development of individual sub-systems. Hence, the notation used in connection with the longitudinal-transverse subsysteming may be used to describe the genetic code taking into accounts the sub-systems [8, 17-19].

Sub-systems described in this way are embedded, therefore, when we describe each of the sub-systems immersed at any depth of subsysteming, the relation between the whole and a given part may exist, and there may be different system situations in which the cross-over and mutation operator may appear.

## 1.2. System operator of longitudinal-transverse cross-over

The main title (on the first page) should begin 1 3/16 inches (7 picas) from the top edge of the page, centered, and in Times New Roman 13-point, boldface type. Capitalize the first letter of nouns, pronouns, verbs, adjectives, and adverbs; do not capitalize articles, coordinate conjunctions, or prepositions (unless the title begins with such a word). Please initially capitalize only the first word in other titles, including section titles and first, second, and third-order headings (for example, “Titles and headings” — as in these guidelines). Leave two blank lines after the title.

The operator of longitudinal-transverse cross-over is the operator exchanging the sub-systems, systems, elements, parameters, etc. between the systems, which undergo the same operations of longitudinal-transverse subsysteming. As a result of the exchange of parts of the system between the two parent-systems two new systems (two descendants) are obtained, being new solutions to the problems.

The operation initiated by the cross-over operator consists of three stages: first systems belonging to the initial population, ordered by the value of the fitness (adaptation) function, are paired, then the longitudinal-transverse subsysteming operation is performed on the pairs of systems. Finally, parts of the two systems that form a pair (genetic material that constitutes fragments of code sequences between parent systems) are exchanged. The cross-over operation takes place with a certain probability called cross-over probability $p_k$.

Due to different depths of longitudinal-transverse subsysteming, different methods of system longitudinal-transverse cross-over may be distinguished, e.g.: 1st level longitudinal cross-over, 1st level longitudinal-transverse cross-over, 2nd level longitudinal cross-over, etc.

**Table 1.** Structure of input and output parameters of electric power system

| θ | $u_1$ | $u_2$ | $u_3$ | $u_4$ | $u_5$ | $u_6$ | $u_7$ | $u_8$ | $u_9$ | $u_{10}$ | $u_{11}$ | $u_{12}$ | $u_{13}$ | $u_{14}$ | $y_1$ |
|---|---|---|---|---|---|---|---|---|---|---|---|---|---|---|---|
| 1946 | 3000 | 2553 | 174 | 333 | 3224 | 24420 | 882 | 76000 | 7091 | 3550 | 190 | 15 | 176 | 21 | 2004 |
| … | …. | … | … | … | … | … | … | … | … | … | … | … | … | … | … |
| 2007 | 35096 | 35800 | 254 | 591 | 243000 | 359526 | 49500 | 759500 | 201500 | 45150 | 60200 | 1704800 | 35500 | 7752 | 34877 |

In the 1st level longitudinal cross-over, the process of exchange of the sub-systems, obtained as a result of the action performed by the longitudinal subsysteming operator, involves exchange of places between the operating subsystems and securing subsystems of both systems.

Similarly, in the 1st level transverse cross-over the process of exchange of the sub-systems, obtained as a result of the action performed by the transverse subsysteming operator, involves exchange of places between the control subsystems and realization subsystems of the both systems.

In the 1st level longitudinal-transverse through cross-over, the process of exchange of the subsystems, obtained as a result of the action performed by the longitudinal-transverse subsysteming operator, involves exchange of places between the operating subsystem from the control subsystem of one system and securing system from the realization subsystem of the second system.

Methods of cross-over between systems, which underwent the operation of higher level (2nd, 3rd, etc. level) longitudinal-transverse subsysteming.

Moreover, we may distinguish a multi-point cross-over which takes place in a praxeological progression of systems, according to the same above mentioned method.

## 1.3. System operator of longitudinal-transverse mutation

System operator of longitudinal-transverse operator is the operator that adds new subsystems, systems, elements to the structures of the existing systems (or removing the existing subsystems, systems, elements, or even deforming them), which are subject to the same operations of longitudinal-transverse sub-systeming. As a result of the operation of the mutation operator new systems come into being, and mutation may take different forms:

- 1st level longitudinal mutation,
- 1st level transverse mutation,
- 1st level longitudinal-transverse through mutation,
- 2nd level longitudinal mutation, etc.

Mutation involves the change of the subsystem, system, element into another subsystem, system, element, and, in particular, it involves adding a new subsystem, system, element or removing the corresponding existing part of the system and takes place with the probability $p_m$.

## 1.4. System fitness function as an robustness function

The notion of fitness function was defined in the work [17-27], namely in three various:

$$1)\quad \Delta y = y(t) - avg(y(t)),$$
$$2)\quad \Delta y = y(t) - \max(y(t)), \qquad (2)$$
$$3)\quad \Delta y = y(t) - avg(y(t)).$$

where:

y(t) – value of evaluation function in generation t,

avg(y(t)) – average value of the evaluation function of all individuals in generation t,
max(y(t)) – maximum value of the evaluation function of all individuals in generation t.

## 2. Experiment of the identification of the system

In order to conduct the identification for the purpose of constructing the initial population of the SAE algorithm, measurement data included in the work [17-19], concerning the system of manufacturing of power and electrical energy, were used. Data that characterize the system describe 14 input variables ($u_1$–$u_{14}$) and 4 output variables ($y_1$-$y_4$) in the yearly perspective for the period of 1946-2007, and have the structure that is presented in table I [21-25].

For the purpose of identification the arx method and the model of MISO[3] type was used and 4 models were obtained in this way [7, 14]. They were obtained for individual outputs and all inputs and they constitute the model of the KSE system. Individual models can be written down using the following mathematical formula:

$$A_i(q)y_i(t) = B_i(q)\cdot u_i(t) + e(\theta), \tag{3}$$

where: i=1-4.

Dane concerning individual input and output variables of the KSE system, following the method used in paper [48] were divided into 30 periods with the step equal 1 year, which allows to obtain 33 periods and the same number of progressive models of the KSE system. Therefore, conducting the identification for these periods in the MATLAB environment using System Identification Toolbox, for fourteen input variables and the first output, representing employment in power plants (total), the following model was obtained [17-19]:

$$A(q)\cdot y_1(\theta) = B(q)\cdot u_1(\theta) + e(\theta), \tag{4}$$

where:

$$\begin{gathered} A(q)\cdot y(\theta) = 32585 - 0.1342 * 32698, \\ B_1(q)\cdot u_1(\theta) = -0.1342\cdot 53711 - 0.05387\cdot 54915 - 0.1443\cdot 55371, \end{gathered} \tag{5}$$

and:

$$\begin{gathered} B(q)\cdot u(\theta) = \sum_{i=1}^{14} B_i(q)\cdot u_i(\theta), \\ B_1(q)u_1(\theta) = \sum_{j=1}^{8} B_i(q^{-j})\cdot u_i(\theta - j), \end{gathered} \tag{6}$$

thus:

$q^{-1}$ - time-shift operator,

$q^{-1}\cdot u(\theta) = u(\theta - 1)$,

$A(q) = 1 - 0.1342\cdot q^{-1}$,

$B1(q) = -0.1342q^{-1} - 0.05387q^{-2} - 0.1443\cdot q^{-3}$,

$B2(q) = -0.1965q^{-1} - 0.7748q^{-2} + 0.3264\cdot q^{-3}$,

$B3(q) = -5.191q^{-1} + 0.3683q^{-2} + 29.52\cdot q^{-3}$,

$B4(q) = -14.5q^{-1} + 9.715q^{-2} + 14.17\cdot q^{-3}$,

$B5(q) = 0.1554q^{-1} - 0.05293q^{-2} + 0.06803\cdot q^{-3}$,

$B6(q) = 0.01335q^{-1} - 0.02755q^{-2} + 0.006739q^{-3}$,

$B7(q) = -0.05234q^{-1} - 0.2766q^{-2} - 0.6104\cdot q^{-3}$

[3] MISO – Multiple Input Single Output

$$B8(q) = -0.002718 q^{-1} - 0.007408 q^{-2} + 0.02639 \cdot q^{-3},$$
$$B9(q) = -0.03015 q^{-1} + 0.1073 q^{-2} - 0.02883 \cdot q^{-3}$$
$$B10(q) = 0.08841 q^{-1} + 0.2976 q^{-2} + 0.1541 \cdot q^{-3},$$
$$B11(q) = 0.131 q^{-1} + 0.06231 q^{-2} - 0.04736 \cdot q^{-3},$$
$$B12(q) = 0.01546 q^{-1} - 0.009961 q^{-2} - 0.009608 \cdot q^{-3},$$
$$B13(q) = -0.02857 q^{-1} - 0.1287 q^{-2} + 0.2337 \cdot q^{-3},$$
$$B14(q) = -0.02198 q^{-1} + 0.3019 q^{-2} + 0.02655 \cdot q^{-3}.$$

In order to construct an initial system population a simplified SISO model was used, which has the following form [2, 21-25]:

$$\begin{aligned} y_1(\theta) = {} & 0.1342 \cdot y(\theta-1) - 0.1342 \cdot u_1(\theta-1) - 0.05387 u_1(\theta-2) - 0.1443 u_1(\theta-3) - 0.1965 u_2(\theta-1) - 0.7748 u_2(\theta-2) + \\ & + 0.3264 u_2(\theta-3) - 5.191 \cdot u_3(\theta-1) + 0.3683 u_3(\theta-2) + 29.52 \cdot u_3(\theta-3) - 14.5 \cdot u_4(\theta-1) + 9.715 \cdot u_4(\theta-2) + \\ & + 14.17 u_4(\theta-3) + 0.1554 u_5(\theta-1) - 0.05293 u_5(\theta-2) + 0.06803 u_5(\theta-3) + 0.01335 u_6(\theta-1) - 0.02755 u_6(\theta-2) + \\ & + 0.006739 u_6(\theta-3) - 0.05234 u_7(\theta-1) - 0.2766 u_7(\theta-2) - 0.6104 u_7(\theta-3) - 0.002718 u_8(\theta-1) - 0.007408 u_8(\theta-2) + \\ & + 0.02639 u_8(\theta-3) - 0.03015 u_9(\theta-1) + 0.1073 u_9(\theta-2) - 0.02883 u_9(\theta-3) + 0.08841 u_{10}(\theta-1) + 0.2976 u_{10}(\theta-2) + \\ & + 0.1541 u_{10}(\theta-3) + 0.131 \cdot u_{11}(\theta-1) + 0.06231 u_{11}(\theta-2) - 0.04736 u_{11}(\theta-3) + 0.01546 u_{12}(\theta-1) - 0.009961 u_{12}(\theta-2) + \\ & - 0.009608 u_{12}(\theta-3) + -0.02857 u_{13}(\theta-1) - 0.1287 \cdot u_{13}(\theta-2) + 0.2337 \cdot u_{13}(\theta-3) - 0.02198 u_{14}(\theta-1) + 0.3019 u_{14}(\theta-2) + \\ & + 0.02655 u_{14}(\theta-3) + e(\theta). \end{aligned} \tag{7}$$

where:

$y_1$ (θ)- achievable power in power plants (total) [MW],

$u_1$ (θ)- employment in power plants (total) [persons], etc.

After conducting the identification and obtaining the results in the form of the model of the system, an initial system population was built with the following structure [2, 10, 17-19]:

$$ch = [ch_a \quad ch_b] \tag{8}$$

where:

$$\begin{aligned} ch_a &= [a_1 \cdot q^{-1} \quad a_2 \cdot q^{-2} \quad a_3 \cdot q^{-3} \quad a_4 \cdot q^{-4} \quad a_5 \cdot q^{-5}], \\ ch_b &= [b_1 \cdot q^{-1} \quad b_1 \cdot q^{-2} \quad b_1 \cdot q^{-3} \quad b_1 \cdot q^{-4} \quad b_1 \cdot q^{-5} \quad b_1 \cdot q^{-6} \quad b_1 \cdot q^{-7} \quad b_1 \cdot q^{-8}]. \end{aligned} \tag{9}$$

After substituting numerical data, an initial population that consists of chromosomes that have the structure of base individuals is obtained:

$$\begin{aligned} ch_1 &= [[-0{,}1342 \quad 0 \quad 0 \quad 0 \quad 0] \quad [0{,}343 \quad -0{,}05387 \quad -0{,}1443 \quad 0 \quad 0 \quad 0 \quad 0 \quad 0]] \\ & \ldots\ldots\ldots\ldots\ldots\ldots\ldots\ldots\ldots\ldots\ldots\ldots\ldots\ldots\ldots\ldots\ldots\ldots\ldots\ldots \\ ch_{33} &= [[-0{,}7413 \quad 0{,}07914 \quad 0{,}04467 \quad 0 -17{,}04 \quad 1{,}427] \quad [0 \quad 0 \quad 0 \quad 0 \quad 0 \quad 0 \quad 0 \quad 0]] \end{aligned} \tag{10}$$

In order to increase the number of individuals in the population to be equal 99, subsequent chromosomes were generated on the basis of base individuals, taking into account the limits of divergence and the precision of the model obtained as a result of identification. Two new individuals are formed from one base individual, and the genes of the child (derivative) individual are generated on the basis of appropriate genes possessed by the base individual, and their value is a random variable from the range <value of base gene - *range;* value of base gene + *range*>, where the variable range is equal to 0.001 (Table 2) [2].

In the general case the following is obtained for arx model:

$$A(q)\cdot y(\theta)=B(q)\cdot u(\theta), \tag{11}$$

where:
u(θ) – observation of the object input in a discrete step θ,
y(θ) – observation of the object output in a discrete step θ,
$A(q)=1 + a_1 q^{-1} + a_2q^{-2} + \dots + a_{na}q^{-na}$,
$B(q)=b_0 + b_1q^{-1} + b_2q^{-2} + \dots + b_{nb}q^{-nb}$,
na – the degree of the polynomial A(q),
nb - the degree of the polynomial B(q),
nk – delay between the output and the input.

**Table 2.** Variants of the algorithm depending on the genetic operators and the robustness function

| **Variant of algorithm** | **Selection** | | | **Cross-over** | | | **Mutation** | | | **Adaptation function** | | |
|---|---|---|---|---|---|---|---|---|---|---|---|---|
| | **Roulette** | **Tourna ment** | **Ranking** | **Single-point** | **Double-point** | **Uniform** | **Single-point** | **Multi-point** | **Probabi listic** | **I** | **II** | **III** |
| **I.I** | √ | | | √ | | | √ | | | √ | | |
| **I.II** | | √ | | | √ | | | √ | | √ | | |
| **I.III** | | | √ | | | √ | | | √ | √ | | |
| **II.I** | √ | | | √ | | | √ | | | | √ | |
| **II.II** | | √ | | | √ | | | √ | | | √ | |
| **II.III** | | | √ | | | √ | | | √ | | √ | |
| **III.I** | √ | | | √ | | | √ | | | | | √ |
| **III.II** | | √ | | | √ | | | √ | | | | √ |
| **III.III** | | | √ | | | √ | | | √ | | | √ |

Model (11) may be written down in the form of a differential equation, as follows [2, 8, 17-19]:

$$y(\theta)+\dots+a_{na}\cdot y(\theta-na)=b_0\cdot u(\theta-nk)+b_1\cdot u(\theta-nk-1)+\dots+b_{nb}\cdot u(\theta-nk-nb). \tag{12}$$

Assuming further that θ=n+1, forecast values of the output variables may be determined on the basis of the model obtained as a result of identification:

$$y(n+1)=-a_n\cdot y(n)-a_{n-1}\cdot y(n-1)-\dots-a_{na}\cdot y(n-na)+b_0\cdot u(n-nk)+\dots+b_{nb}\cdot u(n-nk-nb). \tag{13}$$

For the chromosome generated:

$$ch_1=[[-0{,}1342\quad 0\quad 0\quad 0\quad 0]\quad [0{,}343\quad -0{,}05387\quad -0{,}1443\quad 0\quad 0\quad 0\quad 0\quad 0]]. \tag{14}$$

For the discussed model of the subsystem of power plants system (specific values) function (13) assumes the following form:

$$y(n+1)=0.1243\cdot y(n)+0.343\cdot u(n)-0.05387\cdot u(n-1)-0.1443\cdot u(n-2), \tag{15}$$

## 3. Implementation of the SAE algorithm

The SAE evolutionary algorithm was implemented using three programming languages: Matlab, $C^{++}$ , Java and in three variants concerning genetic operators and in three variants concerning the adaptation function [2,11,17-27]. As a result of the functioning of the evolutionary algorithm, a fit model of the subsystem of power and electrical power production was obtained in respect of the adaptation function (variant 1).

The model has the following form:

$$ch = [ch_1 \quad ch_2]$$
$$ch_1 = [-0{,}7419 \quad 2{,}9478E-4 \quad 5{,}8129E-5 \quad -17{,}0401 \quad 1{,}0565E-4\ ] \qquad (16)$$
$$ch_2 = [0{,}3428 \quad 0{,}6339 \quad 3{,}0696e-4 \quad -8{,}0137e-4 \quad 0{,}0951 \quad 0{,}2837 \quad 4{,}2003e-4 \quad 7.5713e-4]$$

Due to the fact that the obtained model only applied to the impact of employment in power plants, as well as the impact of technical equipment in the workplace, automation and robotization, on the level of total achievable power in all professional power plants in Poland, the interpretation of individual elements of the obtained model [2,10,17-19] is next:

$a_1\left[\frac{kW}{kW}\right]$ - the ratio of power to the power produced in the year (t-1),

.....................................................................

$a_5\left[\frac{kW}{kW}\right]$ - the ratio of power to the power produced in the year (t-5),

$b_1\left[\frac{kW}{person}\right]$ - the ratio of power produced in year t to employment (t-1),

........................................................................

$b_8\left[\frac{kW}{person}\right]$ - the ratio of power produced in year t to employment (t-8).

# 4. Examining the fitness of the algorithm SAE

## 4.1. Description of the conducted examinations

Type your main text in 10-point Times New Roman, single-spaced with 10-point interline spacing. Do not use double-spacing. All paragraphs should be indented 1 pica (approximately 1/6- or 0.17-inch or 0.422 cm). Be sure your text is fully justified—that is, flush left and flush right. Please do not place any additional blank lines between paragraphs.

The conducted examinations involved the execution of the SAE algorithm implemented in three programming languages. Different elements of the algorithm, e.g. adaptation (robustness) function, the number of generations, the cost of execution of the algorithm measured in milliseconds and the probability of mutation were taken into account. The results of the examination were presented in three kings of charts [2]:

- the ratio of the fitness function to the number of generations,
- the ration of the cost of execution of the algorithm to the number of generations,
- the ratio of the fitness function to the probability of mutation.

The charts present the fitness of the algorithm, divided by variants. The following was assumed:

- probability of cross-over – 0.75, probability of mutation – 0.01, number of generations – 1000.

Each of the languages used for the implementation (Java, C++, Matlab) was evaluated using the scale 0-2 for the sake of clarity of the evaluation (2 points were given for the best result in a given category, and 0 points were given for the worst result). The points were awarded in three categories: the best final result, the lowest cost, the lowest susceptibility.

Variant I.I of the algorithm means that the following genetic operators were implemented: roulette circle selection, single-point cross-over, single-point mutation (table 3).

In addition, in variant I the following form of the adaptation function was used:

$$\Delta y = y(t) - avg(y(t)) \qquad (17)$$

**Table 2.** Ranking of programming languages – variant II

| **Programming language** | **The best result** | **The lowest cost** | **The lowest susceptibility** |
|---|---|---|---|
| **Java** | 2 | 1 | 2 |
| **C++** | 1 | 2 | 0 |
| **MATLAB** | 0 | 0 | 1 |

The best result was obtained using the algorithm implemented in Java language, and the worst result was obtained using the algorithm implemented in Matlab. The shortest time of execution was recorded for the program written in C++, and the execution of program in Matlab took the longest. Depending on the value of probability of mutation, the version of the algorithm implemented in Java is the least susceptible to the changes of this parameter and, and the implementation based on $C^{++}$ is the most susceptible. The optimum value of the probability of mutation is 0.01.

### 4.2. Final evaluation of the operation of the SAE algorithm

The final ranking is presented in table 4. All rankings considered, the optimum language seems $C^{++}$ as it is the fastest as regards calculations and computation. However, each of the languages has its own forte. Java provides the best results, C++ is the fastest, and Matlab is the least susceptible to the changes of the probability of mutation. It is worth emphasizing that although Java provides the best results, it is on the 2nd place in the susceptibility ranking, and the cost is a little higher than in the C++ implementation. In consequence, Java is only a little weaker than C++ in this respect although it has the highest position in the ranking.

It is also important to select the variant of the algorithm, which could be used to conduct the research on the development of the system. The optimum variant of the algorithm should be as stable as possible and return the best possible results. It is difficult to select one best implementation from the nine that were tested, as the results achieved by each variant I.I, I.II, I.III, II.I, II.II and II.III were very similar. Variants III.I, III.II and III.III should not be taken into consideration as the results achieved by the variants differ significantly from other variants. It is not caused by adding one gene to the genotype, but it is rather due to its value, which was randomly selected from the range of values limited by the smallest and the largest value of all the genes [2].

**Table 4.** Final ranking of programming languages

| Programming language | The best result | The lowest cost | The lowest susceptibility |
|---|---|---|---|
| Java | 2 | 1 | 2 |
| C++ | 1 | 2 | 0 |
| MATLAB | 0 | 0 | 1 |

## 5. Conclusion

The paper attempts to define a new genetic algorithm, which was called a System Evolutionary Algorithm SAE, which was verified in different practical situations on the basis of the selected subsystem of the domestic electrical energy system KSE as regards evolving (developing) systems. Further direction of development and research may be finding and defining a better fitness function of the algorithm as well as finding and determining the balance between efficiency and effectiveness of development in this function. Further genetic operators might be implemented for that purpose.

The implementation of systemically Evolutionary Algorithm in Java, $C^{++}$ and Matlab showed that the most appropriate environment is Java, although some criteria for such an environment can be Matalab environment such as $C^{++}$.